\documentclass{svproc}
\usepackage{url}
\usepackage{graphicx}
\usepackage{amsfonts}
\usepackage{amsmath}%
\usepackage{amssymb}
\usepackage{booktabs}
\newcommand{\E}{{\mathbb{E}}}

\begin{document}
\mainmatter              % start of a contribution
\title{SetGAN: Improving the stability and diversity of generative models through a permutation invariant architecture}
\titlerunning{SetGAN}  % abbreviated title (for running head)
\author{Alessandro Ferrero \and Shireen Elhabian \and Ross Whitaker}
\authorrunning{Alessandro Ferrero et al.} % abbreviated author list (for running head)
\tocauthor{Alessandro Ferrero, Shireen Elhabian, Ross Whitaker}
\institute{Scientific Computing and Imaging Institute, University of Utah\\
72 S Central Campus Drive, Room 3750, Salt Lake City, UT 84112, USA\\
\email{alessandro.ferrero@utah.edu} \\
\email{shireen@sci.utah.edu} \\
\email{whitaker@cs.utah.edu}}

\maketitle              % typeset the title of the contribution

\begin{abstract}
Generative adversarial networks (GANs) have proven effective in modeling high-dimensional distributions.  However, their training instability is a well-known hindrance to convergence, which results in practical challenges in their applications. Furthermore, even when convergence is reached, GANs can be affected by mode collapse, a phenomenon for which the generator learns to model only a small part of the target distribution, disregarding the vast majority of the data manifold or distribution. This paper addresses these challenges by introducing SetGAN, an adversarial architecture that processes sets of generated and real samples, and discriminates between the origins of these sets (i.e., training versus generated data) in a flexible, permutation invariant manner. The advantages of this approach are studied theoretically while the state-of-the-art evaluation methods show that the proposed architecture, when compared with GAN variants stemming from similar strategies, produces more accurate models of the input data with lower sensitivity to hyper-parameter settings.
% We would like to encourage you to list your keywords within
% the abstract section using the \keywords{...} command.
\end{abstract}
\section{Introduction}
Since their introduction by \cite{goodfellow14}, {\em generative adversarial networks} (GANs)  have proven to be an effective tool for representing or modeling high-dimensional, complex probability distributions from a set of i.i.d. training samples. Among others, \cite{Karras2017ProgressiveGO} and \cite{brock2018large} showed that impressive results can be achieved, but some challenges remain in training GANs. Common problems are the instability that GANs experience during training and {\em mode collapse}: the generator often produces samples that are far from the target distribution, or can fool the discriminator by modeling a very small set of high-probability samples from the data distribution (e.g., one or more modes). The right choice of hyper-parameters can solve this problem, but the optimal ranges of values are sometimes very narrow and difficult to identify.\\
\indent In recent years, several training strategies have been proposed to mitigate these issues. \cite{Metz2016UnrolledGA}, for instance, modified the training strategy, such that each generator update at time $t$ is computed based on the discriminator's state at time $t+n$ to match the gradients that the optimal discriminator would allow. This strategy stabilizes GAN training and decreases mode collapse, but it significantly increases the computational complexity of each update step.\\
\indent Other strategies modify the GAN's loss function and network architecture. For instance, the loss proposed by \cite{pmlr-v70-arjovsky17a} minimizes a surrogate of the Wasserstein distance between distributions, rather than the Jensen-Shannon divergence, in order to avoid the vanishing of gradients during training. \cite{ZhaoML16} proposed a nonstochastic alternative loss and an autoencoder-like architecture for the discriminator to learn the manifold of the training data.\\ 
\indent  \cite{Salimans2016ImprovedTF} and \cite{NIPS2018_7350} tackled the problem with a different approach, providing the discriminator with information from sets of samples rather than single samples. In the {\em minibatch discrimination} (MD), the discriminator calculates an indicator of distance between each example and the samples from the same distribution in a minibatch, using this information, along with the learned latent representation, to individually classify each sample. The additional information helps to avoid mode collapse, but this method looses the guaranteed optimality properties of the original GAN formulation. Following the philosophy of minibatch discrimination, \cite{NIPS2018_7423} designed PacGAN, a technique that stacks examples from the same distribution in a pack and feeds these stacks to the discriminator. 
Hence, the \textit{one} classification that each pack receives from the discriminator is based on the collective information provided by samples within the pack. A GAN whose discrimination task is performed on packs of samples, rather than on single examples, is more stable during training and more resistant to mode collapse. 
However, since the examples are simply stacked together in the data space, the functional mapping $D$ learned by the discriminator is not permutation invariant, i.e., the order in which the samples are stacked in the pack affects the outcome of $D$. Theoretically, this architecture can learn to be invariant to the samples' order, but, as shown in this paper's empirical results, the quality of the generated distribution is degraded.\\
\indent In this paper, we address these challenges by directly modifying the GAN objective function such that it compares the \textit{Cartesian products} of the real and generated distributions. We show that this reduces mode collapse and stabilizes the training phase. In particular, the proposed adversarial architecture, \textit{SetGAN}, relies on equivariant layers (similar to \cite{Zaheer2017DeepS}) that treat individual samples identically, followed by a permutation invariant architecture that produces a rich statistical summary of the feature space of these equivariant layers.
We examine the proposed approach in a more general, stochastic framework, with the understanding that it is complementary to various other existing improvements in GAN training and regularization.

\section{SetGAN Forumlation}

The strategy of the proposed {\em SetGAN} architecture is to allow the discriminator to make a decision on a {\em set} of samples from generator and/or training (i.e., real) data. Let $\mathbf{x}_i \in {\cal X} \subset \mathbb{R}^d$ denote samples from the data space and $\mathbf{z}_i \in \mathbb{R}^l$ denote samples from the latent space of the generator. Let $p_\mathbf{x}$ and $p_\mathbf{z}$ be the probability distributions of the data space and the latent space, respectively, where $k$ independent samples are drawn from each. Let $X^{(k)}$ denote the $k-$ary Cartesian product over the training set, where $X^{(k)} = \mathcal{X} \times \cdots \times \mathcal{X} = \left\{ (\mathbf{x}_1, \cdots, \mathbf{x}_k) | \mathbf{x}_i \in \mathcal{X}, \forall i \in \{1, \cdots, k\}\right\}$. In a similar fashion, let $Z^{(k)} = \left\{ (\mathbf{z}_1, \cdots, \mathbf{z}_k) | \mathbf{z}_i \sim p_\mathbf{z}(\mathbf{z}), \forall i \in \{1, \cdots, k\} \right\}$ denote the $k-$ary Cartesian product over the generator latent space. We use the notations $p_\mathbf{x}^{(k)}$ and $p_\mathbf{z}^{(k)}$ to signify the probability distributions defined on the Cartesian products $X^{(k)}$ and $Z^{(k)}$, respectively. The SetGAN loss function is therefore formulated as
\begin{eqnarray}
\min_G \max_D V(G, D) = \mathbb{E}_{X^{(k)} \sim p_\mathbf{x}^{(k)}}\left[ \log D\left(\mathbf{x}_1, \ldots, \mathbf{x}_k\right)\right] \nonumber \\
+
\mathbb{E}_{Z^{(k)} \sim p_\mathbf{z}^{(k)}}\left[ \log (1 - D\left(G(\mathbf{z}_1), \ldots, G(\mathbf{z}_k)\right)\right)].
\label{eq:loss}
\end{eqnarray}

The SetGAN formulation has two advantages: (i) it allows the discriminator to easily recognize repeated or very similar examples in the set of $k$ samples, explicitly forcing the generator to increase the entropy of the learned distribution, decreasing mode collapse; and (ii) it maintains the global optimum property in the idealized setting.  If we let $p_G: \mathcal{X} \mapsto \mathbb{R}^+$ be the probability distribution of the generated samples in the data space, we have:

\vspace{0.1in}
\noindent
{\bf Theorem:} The global optimum  (Nash equilibrium) of the objective function defined in (\ref{eq:loss}) is achieved when $p_G =  p_\mathbf{x}$.
\vspace{0.05in}

\noindent
{\em Proof.}  We define $X^{(k)} = (\mathbf{x}_1, \ldots, \mathbf{x}_k)$ as a single sample consisting of a concatenation of i.i.d. samples from $p_\mathbf{x}^{(k)}$.    Likewise, we define a generator $G$ as producing $k$ i.d.d. samples from inputs $Z^{(k)} = (\mathbf{z}_1, \ldots, \mathbf{z}_k)$, with a density in $\mathcal{X}$ of $p_G^{(k)}$, obtaining the following optimization:
\begin{eqnarray}
\nonumber
\min_G \max_D V(G, D) = \mathbb{E}_{X^{(k)} \sim p_\mathbf{x}^{(k)}}[ \log D(X^{(k)})] \\
+ \mathbb{E}_{X^{(k)} \sim p_G^{(k)}}[ \log(1 - D(X^{(k)}))]
\label{eq:loss2}
\end{eqnarray}

From the proof in \cite{goodfellow14},  we have that the global optimum for (\ref{eq:loss2}) is $p_G^{(k)} =  p_\mathbf{x}^{(k)}$, which implies that 
$p_G(\mathbf{x}_1) \times \ldots \times p_G(\mathbf{x}_k) =  
p_\mathbf{x}(\mathbf{x}_1) \times \ldots \times p_\mathbf{x}(\mathbf{x}_k)$.    Marginalizing over all $i = 1, \dots, k$ except where $\mathbf{x}_i = \mathbf{x}_j$, results in  $p_G(\mathbf{x}_j) =  
p_\mathbf{x}(\mathbf{x}_j)$, for all $\mathbf{x}_j \in \mathcal{X}$, and thus $p_G =  p_\mathbf{x}$. $\hfill\blacksquare$

\vspace{0.1in}

The formulation in (\ref{eq:loss}) suggests a modification to the original GAN formulation where one simply extends the discriminator, $D$, to accept and classify a concatenation of the $k$ samples from $\mathcal{X}$.   However, the ordering of samples in $X^{(k)}$ is arbitrary, because the samples are i.i.d.    Therefore, we treat the $k$ independent samples as a set $\{ \mathbf{x}_1, \ldots, \mathbf{x}_k \}$ and rely on an discriminator architecture that is invariant to permutations of the input samples.

\subsection{Mathematical analysis of SetGAN}

The SetGAN formulation decreases generator-adversary overfitting, where discriminator gradients near the training data dominate the updates to the generator, thereby interferring with generalization.    \cite{Thanh-Tung} examine the quantity:
\begin{equation}
\Delta D(x) = \frac{ || D(x) - D(y) || }{x - y}
\label{eq:delta}
\end{equation}
where $D(\cdot)$ is the discriminator, $y$ is a generated sample, and $x$ is a nearby training sample.
If the discriminator is (almost) optimal, the numerator is
bounded from below. Updates to the generator result in
$y \rightarrow x$, causing Eq. \ref{eq:delta}, in the limit,
to approach the gradient magnitude of the discriminator and
simultaneously {\em explode} (become unbounded).   Thus, derivatives
of the discriminator near the training data, which push nearby generated samples toward the training sample, dominate the generator updates, leading to mode collapse.

While that argument relies on a fixed, finite set of generated
samples that evolve toward the training data, in practice, the
generated samples are drawn from a probability distribution $p_G(u)$.  
We now consider the effects of SetGAN when reformulating the gradients' analysis in the case of mode collapse. First, we characterize the behavior of the optimal discriminator in the case where  $p_G(u)$ is a density {\em function}.
\begin{theorem}
	Let $p_G$  be a probability density function defined on $R^d$ and
	let $X = {x_1, \ldots, x_n}$ be a finite set of training samples.  Then
	there exists a classifier for distinguishing $y \sim p_G$ from
	$x \in X$ with an expected error of zero.  
\end{theorem}
In this case, the ideal classifier returns $1$ for training data samples and zero otherwise.  Because $p_G$ is continuous, the point probabilities at the training data are zero, and thus the expected error is zero.  Under these conditions, a generator would be forced to modify $p_G$ to approximate the measure that describes the probability distribution of the training samples.
Here we describe a simple model of this overfitting scenario.  In this
notation, we evaluate a single training sample, but the analysis
applies to any subset of training data.
\begin{definition}
	A distribution $p_G$ exhibits $\alpha$-$\epsilon$ mode
	collapse with respect to a distribution (kernel) $k(\cdot)$ iff it can be
	represented as
	$
	p_G(u) = f(u) + \alpha k_\epsilon (u-x)
	$
	where $f(u) \geq 0$, $k_\epsilon$ is the kernel scaled/normalized to size
	$\epsilon$, and there exists $u \in R^d$ such that $f(u) = 0$ and $k_\epsilon (u-x) > 0$ (i.e. $\alpha$ is maximal).  
\end{definition}
By this definition, the classic case of mode collapse would be where
$\alpha =1$ and $\epsilon = 0$---$p_G$ is a measure centered at $x$, and the generator consistently produces a single sample.

Because updates to the GAN are stochastic, it is informative to
examine the expected value of Eq. \ref{eq:delta}, which serves (by the
mean value theorem) as a lower bound on the gradient magnitude of the discrimitor
$D$, which we assume classifies all training samples correctly.  
\begin{equation}
\E_{p_G} [ \Delta D(x) ]=  \E_{p_G} \left[ \frac{1}{r} \right] =
\int\frac{p_G(u)}{r(u)} du,
\end{equation}
where $r(u)$ is the distance to the training sample.   We would like
to characterize this gradient in the case of $\alpha$-$\epsilon$ mode collapse
with respect to a Gaussian kernel
as $\epsilon\rightarrow 0$.  In this case the expectation of $1/r$ is
dominated by the kernel.  Working in spherical coordinates we have:
\begin{equation}
\int\frac{p_G(u)}{r(u)} du \approx
\int_{{\cal S}_{d-1}} \int_0^\infty \frac{\alpha G_{d, \epsilon}(r)}{r}
r^{d-1} dr d \phi, 
\end{equation}
where ${\cal S}^d$ is the unit sphere in $d$ dimensions (and $d
\phi$ the associated unit of solid angle) and $G_{d,
	\epsilon}(\cdot)$ is the $d$-dimensional Gaussian with standard
deviation $\epsilon$ expressed in spherical coordinates.  This
integral has a closed form solution.  The angular integral is the
area of the unit sphere and the radial integral can be reduced to a
$d-2$ moment of the one-dimensional Gaussian, resulting in:
\begin{equation}
\E_{p_g} [ \Delta D(x) ]   \approx
\frac{\alpha}{\epsilon}     \frac{1}{2^{1/2}}\frac{\Gamma
	\left(\frac{d-1}{2} \right)}{\Gamma \left(\frac{d}{2} \right)}
\approx
\frac{\alpha}{\epsilon}   \left( \frac{1}{d}\right)^{1/2}.
\end{equation}
Here we see that the expected value of the gradient of the
discriminator {\em explodes} at the rate of $\alpha/\epsilon$ as $p_G$ undergoes mode collapse.
In the case of SetGAN, we have $k$ independent samples in $\Re^{d}$ and the
expectation is computed in $\Re^{dk}$:
\begin{equation}
\E_{p_G^k} [ \Delta D(x) ]
\approx
\int_{{\cal S}_{rd-1}} \int_0^\infty \frac{\alpha^k  G_{dk, \epsilon}(R)}{R}
R^{dk-1} dR d \phi, 
\end{equation}
where $R^2 = r_1^2 + \cdots + r_k^2$, $r_i$ is the distance of the
$i$th sample in the generated set to the training sample $x$, and we
have relied on the separability of the Gaussian to express the joint
probability in the form $G_{dk, \epsilon}(R)$.
The closed-form expression for the SetGAN case is:
\begin{equation}
\E_{p_g^k} [ \Delta D(x) ] \approx
\frac{\alpha^k}{\epsilon}     \left( \frac{1}{dk}\right)^{1/2}.
\label{equation_8}
\end{equation}
This gradient grows with $\alpha^k/\epsilon$.  Our hypothesis is that early on in mode
collapse, relatively few samples are near the training data,
$\alpha << 1$, the expected value of the discriminator-gradient
remains small, making the generator less susceptible to the cycle of
large gradients, overfitting, and mode collapse. 

{
	\begin{figure*}
		\includegraphics[width=\textwidth]{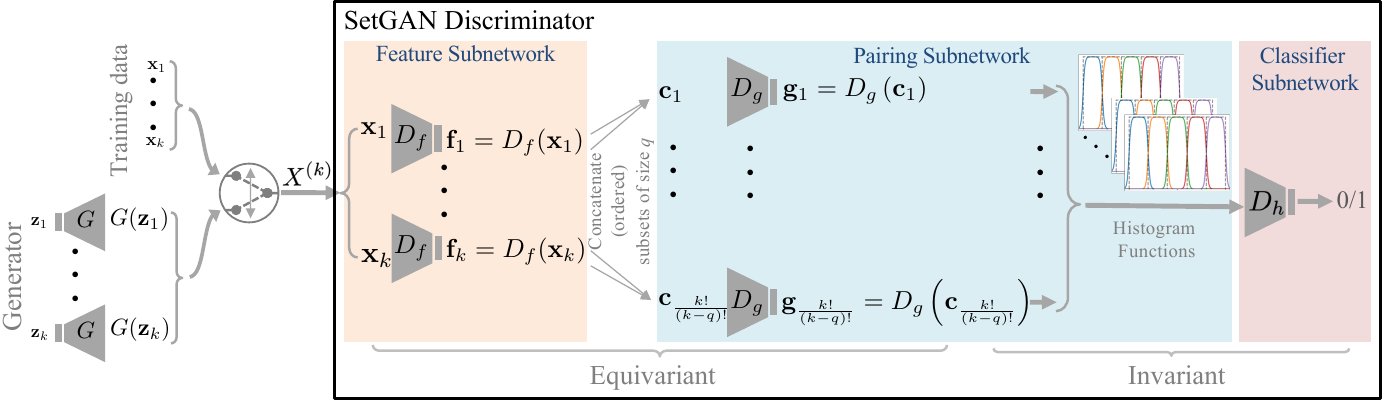}
		\caption{SetGAN architecture}
		\label{fig:setgan}
	\end{figure*}
}

\subsection{Architecture}
Due to its formulation, SetGAN can be implemented into any GAN architecture, regardless of the loss function. In our experiments, we use convolutional generators similar to those proposed by \cite{Salimans2016ImprovedTF}. The discriminator is, however, designed to process sets of samples, invariant to their ordering. SetGAN's discriminator network has three subnetworks: $D_f$, $D_g$, and $D_h$ (see Figure \ref{fig:setgan}). 

The \textit{feature subnetwork} $D_f$ independently maps each example into a latent feature representation. The \textit{pairing subnetwork} $D_g$ processes pairs of feature vectors---where {\em pair} is a permutation of two examples' features from the given set, stacked together in a vector---and learns pairwise relations. The pairing subnetwork uses an \textit{aggregation function} to merge results from each pair, in a permutational invariant manner, right before the \textit{classifier subnetwork} $D_h$.

\indent
In order to provide more informative gradient updates, the aggregation function needs to retain the information coming from each sample in the set. For this task, we construct a differentiable soft histogram approximation, similar to \cite{JonssonBelief}, where the bins' thresholds are represented by logistic functions centered at the bins' boundaries. The subnetwork $D_h$ stacks together the constructed histograms (one for each feature) in a vector that represents the whole set of samples.
$D_g$ processes all possible permutations of two elements in the set to remain invariant to the order of the samples.

\section{Experiments}

In this section, we compare the performance of SetGAN against relevant architectures that mitigate mode collapse in GANs by exposing the discriminator with groups of examples rather than single samples, namely minibatch discrimination (MD) and PacGAN. Taking inspiration from \cite{radford2015unsupervised}, both the generator and the discriminator are implemented as fully convolutional networks with leaky ReLU activations. Following the notation in \cite{NIPS2018_7423}, we define PacGAN5 and SetGAN5 to be the variants of PacGAN and SetGAN whose discriminators classify sets of exactly five samples. The models' evaluation has to encompass different aspects. Following \cite{Heusel2017GANsTB}, the generated samples' quality, is assessed with the Fréchet Inception Distance (FID). In order to quantify mode collapse, the \textit{reverse KL-divergence} is employed. A trained neural network assigns a class to each generated sample, and the resulting modes are the basis for this metric's calculation.

\subsection{Stacked MNIST}
The MNIST dataset by \cite{lecun-mnisthandwrittendigit-2010} contains $70,000$, $28$x$28$ images of hand-written digits. In the stacked MNIST dataset, as described in \cite{Metz2016UnrolledGA}, each sample contains three randomly selected MNIST images. Generating images from this distribution is an  easy  but  not  trivial  problem. Since the amount of modes is high, GAN architectures have the tendency to fall into mode collapse. In this experiment, the latent space dimensionality is high enough to ensure that even the base architecture is able to capture all modes. The $KL$-divergence is computed on $10$ sets, each with $10,000$ generated images, with mean and standard deviation listed in Table \ref{tab:mnist_res}.
\begin{table}[!ht]
	\centering
	\caption{Results for stacked MNIST experiment}
	{%
		\begin{tabular}{|l|c|}
			\hline
			\multicolumn{2}{|c|}{\textbf{Stacked MNIST}}\\
			\hline
			Architecture & Reversed KL-divergence \\
			\hline
			GAN & $0.081 \pm 0.001$ \\
			MD & $0.081\pm  0.002$ \\
			PacGAN5 & $0.258 \pm 0.005$\\
			SetGAN5 & $\mathbf{0.062 \pm 0.001}$ \\
			\hline
	\end{tabular}}\label{tab:mnist_res}
\end{table}

All these GAN variants produce a great variety of recognizable digits with no sign of mode collapse. As expected, GAN is well-suited to learn the stacked MNIST distribution, as shown by the low reversed $KL$-divergence, and MD performs similarly. SetGAN5 generates the closest distribution of modes to the real data, according to the reversed $KL$-divergence. On the contrary, PacGAN5 preforms poorly: this is likely due to PacGAN's need to learn the invariance to the samples' order that SetGAN embeds architecturally. This detail makes a great difference in practice, although theoretically, PacGAN and SetGAN guarantee the same properties.

{
	\begin{figure*}[!ht]
		\includegraphics[width=\textwidth]{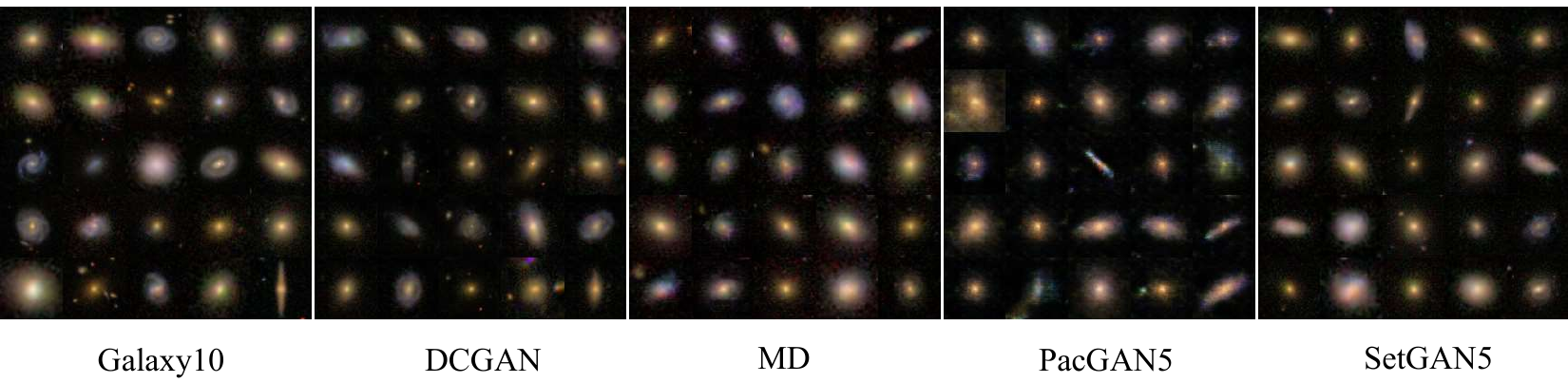}
		\caption{Samples from the Galaxy10 dataset and the generated distributions.}
		\label{fig:astroNN}
	\end{figure*}
}

\subsection{Galaxy10}

The \textit{Galaxy10} dataset is a collection of $21,785$ astronomical, 69x69 photographs of galaxies selected through spectroscopy algorithms by \cite{2002AJ....124.1810S} as part of the Sloan Digital Sky Survey (SDSS). The difficulty in modeling this distribution lies in its class imbalance: some classes are present in less than $1\%$ of the samples, therefore the generator could easily avoid learning these modes and fall into mode collapse. The metrics in Table \ref{tab:galaxy10_celeba_res} show that the base architecture produces high quality samples, while keeping the classes' probabilities close to the real distribution. \\
\indent Neither PacGAN5 or MD improve on the base architecture: PacGAN5's noisy images increase the FID score and both result in a reduced sample quality and degraded distributions, as evidenced by the metrics. SetGAN5, as in the previous experiment, achieves the highest quality of samples. Mode collapse is also decreased and even low probability modes are better represented in SetGAN5 than in any of the other GAN variants, hence the low reversed $KL$-divergence. 

\begin{table}[!ht]
	\centering
	\caption{Quantitative results for Galaxy10 and CelebA experiments}
	\begin{tabular}{|l|cc|c|}
		\hline
		& \multicolumn{2}{c|}{\textbf{Galaxy10}} & \textbf{CelebA}\\
		\hline
		Architecture & FID & Reversed KL-divergence & FID\\
		\hline
		GAN & $25.85 \pm 0.32$ & $0.052 \pm 0.001$ & $25.51 \pm 0.22$\\
		MD & $49.97 \pm 0.28$ & $0.115 \pm  0.004$ & $24.01 \pm 0.26$ \\
		PacGAN5 & $121.10 \pm 0.39$ & $0.207 \pm 0.009$&  $156.61 \pm 0.36$ \\
		SetGAN5 & $\mathbf{22.50 \pm 0.26}$ & $\mathbf{0.040 \pm 0.003}$ & $\mathbf{18.92 \pm 0.25}$ \\
		\hline
	\end{tabular}
	\label{tab:galaxy10_celeba_res}
\end{table}
\vspace{-5mm}

\subsection{CelebA}

The widely-used CelebA dataset, introduced by \cite{liu2015faceattributes}, contains roughly $202,000$ celebrity photographs and is an ideal test for GANs, since the image space is restricted to faces, which still have a wide variability. For this experiment, all images are centered and resized to 64x64 pixels. To examine the sensitivity of each architecture to different hyper-parameters settings, all the architectures were trained multiple times---changing the learning rate and the ratio of iterations between generator and discriminator (we will refer to this as the $GD$ ratio). The tested $GD$ ratio values are 1, 2 and 3, while the learning rate was sampled randomly from the uniform distribution in the range $[0.00002, 0.002]$. For fair comparison, each architecture trained until the model stopped improving its FID score.

\begin{figure*}[!ht]
	\includegraphics[width=1.0\textwidth]{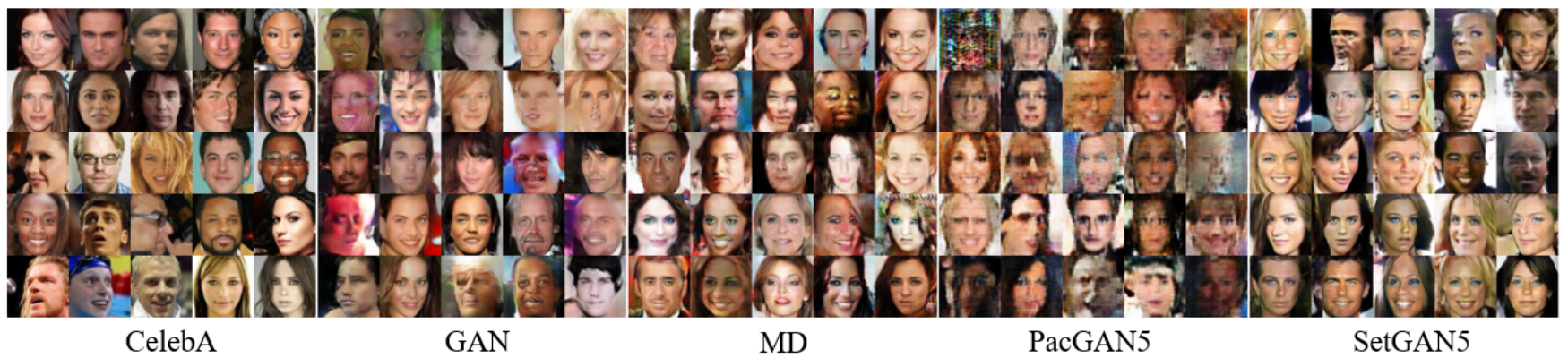}
	\caption{Generated samples for CelebA. SetGAN5 produces sharp images with high variety}
	\label{fig:celebA_samples}
\end{figure*}

{
	\begin{figure*}[!ht]
		\includegraphics[width=1.0\textwidth]{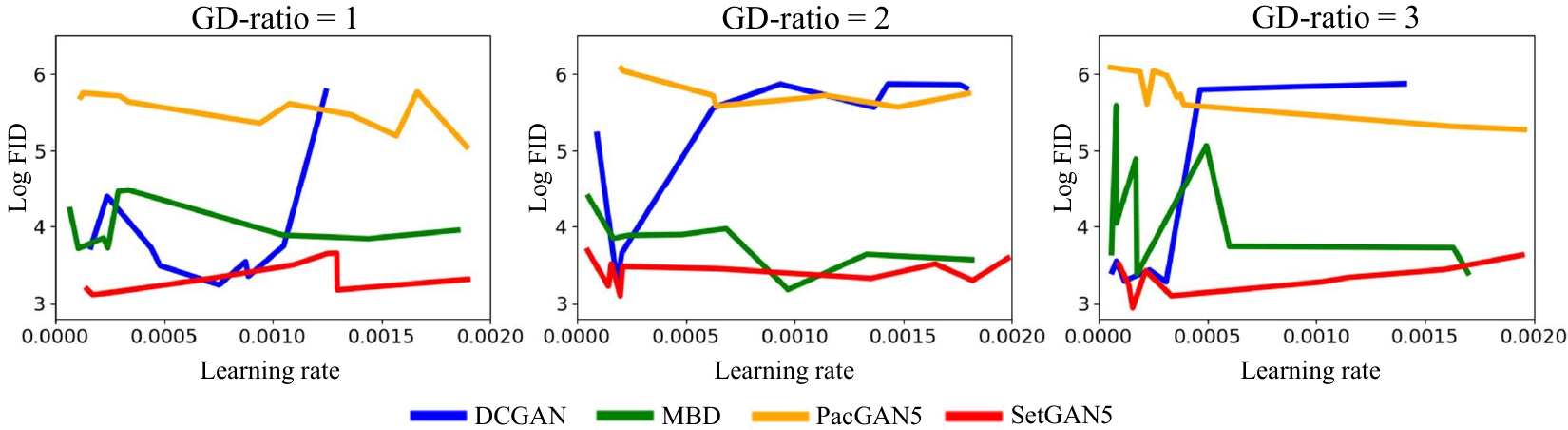}
		\caption{CelebA: analysis on hyper-parameters}
		\label{fig:random_search}
	\end{figure*}
}

Table \ref{tab:galaxy10_celeba_res} reports the best FID score achieved by each architecture given the optimal hyper-parameters and Figure \ref{fig:random_search} shows each architecture's average $\log$ FID scores by learning rate and $GD$ ratio. The Minibatch Discrimination approach improves the image quality when compared to those modeled by the base architecture, as shown by the slight decrease in MB's FID score. PacGAN5 is not able to produce high quality images: although the generator captures general face features, the amount of noise is so elevated that some of the images do not contain faces at all, only noise, and the FID penalizes this deterioration in the samples. SetGAN5 is again the architecture that produces the lowest FID score. SetGAN5 displays flatter graphs than all the others, meaning that the variations in the quality of the generated distribution are comparatively very small, even when the learning rate or the $GD$ ratio change greatly. The Minibatch Discrimination's graph is similar to SetGAN5's, but slightly less stable, especially when the $GD$ ratio is greater than $2$. PacGAN5 and GAN are very sensitive to even small changes in the hyper-parameters, as demonstrated by the high variance in the scores. Furthermore, in many instances, GAN and PacGAN5 suffer from gradient saturation: when the discriminator's loss is too close to $0$, the generator updates are also close to $0$, preventing any learning. This effect is shown by the $\log$ FID values close to $6$. In some instances, the two architectures fall into mode collapse. SetGAN5 and Minibatch Discrimination are not affected by saturation of the gradients or mode collapse in any of the experiments.

\section{Conclusions}
In this paper, we propose a novel architecture, SetGAN, that mitigates the typical failure mode of GANs, i.e., mode collapse, and improves the stability of the training phase. The architecture is less sensitive to hyper-parameter settings, allowing it to avoid saturation of the gradients. 
Additionally, letting the discriminator classify sets of examples, invariant of their order---as opposed to single samples---helps reduce mode collapse by increasing the entropy in the distribution.\\

{\bf Acknowledgements}: This work has been partially supported by the National Institutes of Health [grant number NIAMS-R01AR076120], ExxonMobil, and the Prostate Cancer Foundation Challenge Award. The content is solely the responsibility of the authors and does not represent the official views of the aforementioned entities.


\end{document}